\documentclass[sigconf]{acmart}

\usepackage{multirow}

\AtBeginDocument{%
  }

\acmYear{2023}\copyrightyear{2023}
\acmConference[MobiHoc '23]{The Twenty-fourth International Symposium on Theory, Algorithmic Foundations, and Protocol Design for Mobile Networks and Mobile Computing}{October 23--26, 2023}{Washington, DC, USA (Preprint version)}
\acmBooktitle{The Twenty-fourth International Symposium on Theory, Algorithmic Foundations, and Protocol Design for Mobile Networks and Mobile Computing (MobiHoc '23), October 23--26, 2023, Washington, DC, USA}

\setcopyright{acmcopyright}

\setcopyright{none}
\acmBooktitle{ }
\acmPrice{ }
\acmDOI{ }
\acmISBN{ }

\renewcommand\footnotetextcopyrightpermission[1]{}

\begin{document}

\title[WAFL-Autoencoder]{Detection of Global Anomalies on Distributed IoT Edges \\with Device-to-Device Communication}


\author{Hideya Ochiai}
\affiliation{
  \institution{The University of Tokyo}
  \country{Japan}}
\email{ochiai@elab.ic.i.u-tokyo.ac.jp}
\orcid{0000-0002-4568-6726}

\author{Riku Nishihata}
\affiliation{
  \institution{The University of Tokyo}
  \country{Japan}}
\email{nishihata-riku4792@g.ecc.u-tokyo.ac.jp}

\author{Eisuke Tomiyama}
\affiliation{
  \institution{The University of Tokyo}
  \country{Japan}}
\email{tomiyama-eisuke@g.ecc.u-tokyo.ac.jp}

\author{Yuwei Sun}
\affiliation{
  \institution{The University of Tokyo}
  \country{Japan}}
\email{ywsun@g.ecc.u-tokyo.ac.jp}

\author{Hiroshi Esaki}
\affiliation{
  \institution{The University of Tokyo}
  \country{Japan}}
\email{hiroshi@wide.ad.jp}
\orcid{0000-0001-5657-9216}


\begin{abstract}
Anomaly detection is an important function in IoT applications for finding outliers caused by abnormal events. Anomaly detection sometimes comes with high-frequency data sampling which should be carried out at Edge devices rather than Cloud. In this paper, we consider the case that multiple IoT devices are installed in a single remote site and that they collaboratively detect anomalies from the observations with device-to-device communications. For this, we propose a fully distributed collaborative scheme for training distributed anomaly detectors with Wireless Ad Hoc Federated Learning, namely ``WAFL-Autoencoder''. We introduce the concept of Global Anomaly which sample is not only rare to the local device but rare to all the devices in the target domain. We also propose a distributed threshold-finding algorithm for Global Anomaly detection. With our standard benchmark-based evaluation, we have confirmed that our scheme trained anomaly detectors perfectly across the devices. We have also confirmed that the devices collaboratively found thresholds for Global Anomaly detection with low false positive rates while achieving high true positive rates with few exceptions.
\end{abstract}



\maketitle

\section{Introduction}
Anomaly detection is an important function in Internet of Things (IoT) applications for finding outliers caused by abnormal events such as machine faults, electric surges, or security incidents. Sensors such as cameras, accelerometers, or electric meters can be used for detecting anomalies in remote sites. Here, we must remember that the data generated by these sensors are relatively large because of their sampling frequencies necessary for anomaly detection. 

Uploading all the data to the Cloud is cost-ineffective because most of the data are normal and the normal data consume both Cloud storage and communication bandwidth, increasing the cost of their business. Edge IoT computing is a promising approach for finding anomalies at remote sites from highly-frequent sampling data and for picking up anomaly cases only when they were detected.

In this paper, we consider the case that multiple IoT devices are installed in a single remote site, such as in a building, and that they collaboratively detect anomalies from the observations. They are expected (1) to learn the normal features from daily observations and (2) to detect anomalies when they occurred. 

In such a case, because devices are located physically nearby, we can utilize device-to-device communications for training anomaly detectors. Conventional federated learning \cite{konevcny2016federated,li2020federated} can be also used but all the IoT devices need to exchange model parameters with the Cloud repeatedly, which multiplies the uplink (i.e., cellular) traffic.

We propose WAFL-Autoencoder -- a fully distributed collaborative learning scheme for autoencoder utilizing device-to-device communications. Here, WAFL stands for wireless ad hoc federated learning \cite{ochiai2022wireless}. Even though the name has federated learning, the architecture is fully distributed and different from conventional federated learning. In our previous study, we confirmed WAFL works efficiently for supervised cases. In this paper, we focus on an unsupervised case and study how WAFL allows training for distributed autoencoders. 

\begin{figure}[t]
\centering
\includegraphics[width=0.43\textwidth]{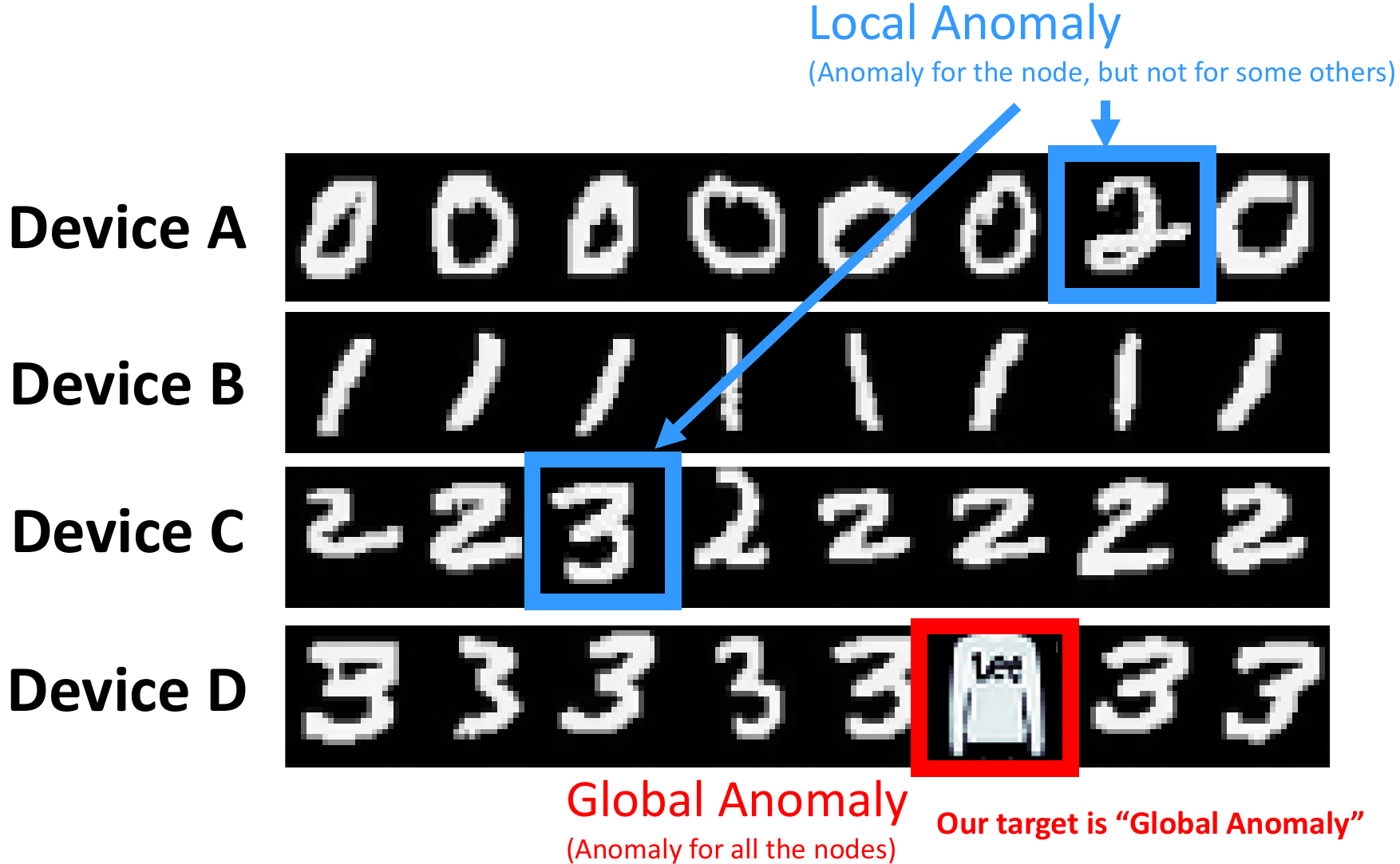}
\caption{Local anomaly and global anomaly examples. Local anomaly is something rare to the device, but well-known to others. Global anomaly is rare to every device, which is difficult to detect with low false-positive rates.}
\label{fig:LocalGlobalAnomaly}
\end{figure}

We also try to use the trained autoencoders for distributed anomaly detection applications. Here, we introduce the concept of Global Anomaly and Local Anomaly as Fig. \ref{fig:LocalGlobalAnomaly}. A local anomaly is something rare to the device, but familiar to others. For example, as Fig. \ref{fig:LocalGlobalAnomaly}, in device A, the major data are handwritten character 0, and the handwritten character 2 is a local anomaly because the 2 is rare to device A, but a major sample at another node. The local anomaly can be easily detected without collaborating with other devices. 

Please note that the shirt image in device D is an abnormal image for all the devices. This is what we call Global Anomaly. Detection of a global anomaly is challenging because the ML model needs to know that the event is also rare to others without sharing the data itself. We focus on the detection of global anomalies as an application of WAFL-Autoencoder. This raises another technical challenge of threshold finding.

A threshold for anomaly detection is usually calculated from the validation data that the device has. However, the calculated threshold in this way is usually skewed for the device's local data. For this problem, we also propose a threshold-finding algorithm in this paper.

Because this is a work-in-progress paper, we focus on understanding the novel characteristics of WAFL-Autoencoder with standard benchmark datasets. The evaluation includes the capability of reconstruction of input samples and the performances of global anomaly detection with pure and dirty cases.

In summary,

\begin{itemize}

\item We propose a WAFL-Autoencoder that utilizes device-to-device communication for learning the features of distributed data samples at IoT edge devices.

\item We introduce the concept of Global Anomaly and Local Anomaly for distributed cases, and focus on Global Anomaly detection, which is a challenging topic.

\item We propose a distributed threshold-finding algorithm for finding Global Anomaly, which should be carried out along with the training of WAFL-Autoencoder.

\item This paper focuses on benchmark-based evaluation with standard MNIST families for understanding the novel characteristics of WAFL-Autoencoder and Global Anomaly detection.

\end{itemize}

This paper is organized as follows. In section 2, we describe our related work. In section 3, we propose WAFL-Autoencoder and threshold finding algorithm, introducing the concept of global anomaly. In section 4, we provide our evaluation. We conclude this paper in Section 5.

\section{Related Work}

There are two types of federated learning: i.e., centralized federated learning (CFL) and decentralized federated learning (DFL). CFL is a well-known architecture, which has a parameter server for model sharing and aggregation. DFL is a new architecture that does not have any central services.

In terms of anomaly detection, we can find some CFL-based studies in computer networks \cite{pei2022personalized, zhao2019multi}, IoT applications \cite{sater2021federated, weinger2022enhancing, cui2021security}, and so forth. Within this architecture, variational autoencoder \cite{zhang2021federated, palato2021federated}, GAN \cite{tabassum2022fedgan, akcay2018ganomaly}, One-Class SVM \cite{anaissi2022personalized} were applied and studied. These systems would be useful if we can set up a centralized server by ourselves, or if we can trust the third-party service provider. However, they still consume the global communication traffic for repeated model exchanges.

The architecture of DFL is studied with several peer-to-peer system architectures. Blockchain-based systems \cite{li2020blockchain, bao2019flchain, pokhrel2020federated, ur2020towards, warnat2021swarm} were proposed for updating global models without any centralized mechanisms. \cite{hu2019decentralized, pappas2021ipls, roy2019braintorrent} were proposed with other forms of peer-to-peer systems.
However, those studies assume the Internet connection, where all the devices need to know remote global addresses with each other. They do not investigate global anomaly detection, which is an important topic of this paper. Besides, usually peer-to-peer systems increase global network traffic. 

Wireless ad hoc federated learning (WAFL) focuses more on opportunistic and ad hoc scenarios where the communication among the devices happens only when they are within the radio range\cite{ochiai2022wireless}. The idea is based on communication without infrastructure, which is named ``wireless ad hoc network (WANET)'' \cite{frodigh2000wireless}. WANET is one of the famous domains in computer network research. The family of WANET include mobile ad hoc network \cite{abolhasan2004review}, vehicular ad hoc network \cite{hartenstein2008tutorial}, and delay/disruption tolerant network \cite{fall2003delay}. They allow communications among today's smart devices with device-to-device communication, i.e., without third-party intervention. The legacy ad hoc networks was for communication not for collaborative machine learning as WAFL proposed.

In this paper, for finding global anomalies in a distributed scenario, we have taken an approach of distributed autoencoders that learn the features collaboratively from distributed sources, which is new in these research fields. We have previously studied the basic learning paradigm in \cite{ochiai2022wireless}, and published the application to distributed generative adversarial network\cite{tomiyama2023gan}. This paper focus on autoencoder for learning the features and finding anomaly in such a distributed scenario.

\section{WAFL-Autoencoder for Anomaly Detection}

To learn the features of data in a distributed environment, we take the approach of distributed autoencoders with distributed anomaly thresholds. 

\subsection{Definition of Terms}

We assume a set of mobile devices $N$ that collaboratively train models based on their local data. Let $W^{n}$ be the model parameters of device $n \in N$, and $X^{n}$ be a set of data samples in $n$. With function $AE$, the output of autoencoder $\hat{x}$ to an input $x \in X^{n}$ at device $n$, can be formulated as:
\begin{equation}
\hat{x}=AE(x,W^{n})
\label{eq:model_prediction}
\end{equation}

With appropriate loss function, such as mean squared error, i.e., $MSE(\hat{x},x)$, $W^{n}$ can be trained to fit specifically for $X^{n}$.

\subsection{Global Anomaly}

In the case of distributed environment, we have at least two types of anomalies as we discussed: Local Anomaly and Global Anomaly. A local anomaly is a rare data sample to the device. It might be important to the device, but it is still ``legitimate'' in general. A global anomaly is rare to everyone, which is more important to be detected. Thus, we focus on the detection of global anomalies in this paper.

\subsubsection{Implicit Data Category and Class}

To formulate a global anomaly, we associate data $x$ with implicit data category and class. We consider that data sample $x$ is implicitly associated with a class $y$. In other words, $x$ could be paired with $y$ as ($x$,$y$). Here, class $y$ belongs to a category denoted by $Y$: $y \in Y$. In this paper, we consider category $Y_0$, e.g., MNIST, as a world of the legitimate dataset. Other
categories such as $Y_1$ and $Y_2$, e.g., Noisy-MNIST and Occluded-MNIST respectively, are the global anomaly domains.

Please note that $y$ and $Y$ are given implicitly and not used for training autoencoder, but we consider there can be such data structure behind $X$. Only the collection of $x$ is given as a dataset for training, and they are not explicitly associated with classes or categories in practice.

\subsubsection{Distributions of Train, Validation, and Test Data}

It is common to assume three types of data: i.e., (1) train, (2) validation, and (3) test data. In this study, we use train data for training the model parameters of the autoencoder. We use validation data for finding a threshold for anomaly detection. We use test data for evaluating the performances of the developed models and anomaly detection. Thus, train and validation data are possessed in the device's local storage, but test data is not stored as an assumption. 

We consider two cases for local data: i.e., pure and dirty cases. In the pure case, we consider that all the train and validation data are legitimate, i.e., implicitly associated with category $Y_0$ only. In the dirty case, we consider that they also contain a small number of global anomaly samples from $Y_i (i \neq 0$). 

Test data should be derived from both legitimate category $Y_0$ and anomaly categories $Y_i ( i \neq 0 )$.

\subsection{Aggregation of Distributed Models with Device-to-Device Communication}

Let $nbr(n)$ be a set of neighbors of device $n$ at a certain time. Here, $nbr(n)$ does not include itself, i.e., $n \notin nbr(n)$. $nbr(n)$ may dynamically change based on the physical mobility of the devices. Device $n$ can directly communicate with all of $nbr(n)$ over wireless channels such as Wi-Fi (ad hoc mode) or Bluetooth. 

Device $n$ is expected to receive all the model parameters from $nbr(n)$. Then, device $n$ aggregates the models with the local model $W^{n}$ by the following formula,
\begin{equation}
W^{n} \leftarrow W^{n} + \lambda \frac{\sum_{k \in nbr(n)}{(W^{k} - W^{n})}}{\vert nbr(n) \vert +1}.
\label{eq:model_aggregation}
\end{equation}

Here, $\lambda$ is a coefficient parameter for adding the differences between the neighbors. The aggregated model needs additional mini-batch-based training with its local data. 

After interacting with many devices, $W^{n}$ becomes generalized and it will be able to correctly predict the data samples that device $n$ does not have but others have. Please note that during this process, WAFL-AE achieves model training without relying on any centralized mechanisms or exchanging the local data itself.

\subsection{Finding Anomaly Threshold with Device-to-Device Communication}

Because of the Non-IID characteristics of distributed environment, a calculated anomaly threshold from local validation data is also bound to the distributions of the local dataset. The thresholds of anomaly scores calculated in this way can be different by devices. For this issue, we propose aggregation of those calculated thresholds to find the proper threshold. This improves the performance of anomaly detection in a distributed environment.

Let $s(x,W^{n})$ be an anomaly score of sample $x$ with model parameter $W^{n}$ at device $n$. According to the work \cite{martin2021fault}, there are many design choices for anomaly score $s$. As an example, we use the following anomaly score $s$ throughout the paper:
\begin{equation}
s(x,W^{n})=\frac{MSE(AE(x,W^{n}),x)}{density(x)},
\end{equation}
\begin{equation}
density(x) = \frac{\sum_{i \in x} p_i}{\vert x \vert}.
\end{equation}

Here, $i$ is the index of a feature of $x$. $p_i \in [0,1]$ is the normalized value of the feature. $\vert x \vert$ is the number of features.  This anomaly score is based on the common MSE-based method with normalization of the input sample. 

With a threshold of anomaly detection $\alpha^{n}$, device $n$ identifies that $x$ is anomaly if $s(x,W^{n}) > \alpha^{n}$, otherwise $x$ is not anomaly. If $W^{n}$ is generalized enough for legitimate samples over the distributed data, the detected anomaly is a global anomaly.

To find proper $\alpha^{n}$, we propose to exchange and aggregate $\alpha^{n}$ among the encounter devices during the model training phase. This can be formulated as follows:  \begin{equation}
\alpha^{n} \leftarrow \frac{ \sum_{k \in nbr(n)}{\alpha^{k} + \alpha^{n} + \gamma \beta^{n}}}{ \vert nbr(n) \vert +1 + \gamma }.
\label{eq:threshold_aggregation}
\end{equation}

This calculates the average of $\alpha$ among the devices encountered. $\beta^{n}$ -- a locally calculated threshold with local validation data at device $n$ is also introduced for averaging (with coefficient $\gamma$).

We calculate $\beta^{n}$ at every time after the model update in the following manner. Given a threshold rate $\delta \in [0,1]$, for example, $\delta=0.999$, the device finds such $s(x,W^{n})$ which cumulative distribution function (CDF) reaches $\delta$ as $\beta^{n}$, i.e., $\beta^{n} = CDF^{-1} \left( \delta; s(X^{n}_\text{val},W^{n}) \right)$. Here, $X^{n}_\text{val}$ is the validation dataset of device $n$.

\section{Evaluation}

To understand the basic characteristics of the WAFL-Autoencoder, i.e., the ability to learn the features, reconstruct legitimate data, and detect global anomalies, we conducted a benchmark-based evaluation as follows.

\subsection{Experiment Setting}

For training, we have configured 10 devices ($n=0, 1, \ldots, 9$) to have 99.95\% Non-IID MNIST dataset for train and validation. MNIST data samples were scattered so that each label $n$ sample was assigned to device $n$ with a probability of 99.95\% and other devices uniformly for the rest of 0.05\%. The ratio between train and validation was 4:1. 

For testing, we have used the original MNIST test dataset, which is IID, as legitimate test data. To evaluate the detection of global anomalies, we have prepared four types of test datasets: (1) Noisy-MNIST, (2) Occluded-MNIST, (3) Fashion-MNIST, and (4) Kuzushiji-MNIST \cite{clanuwat2018deep}. We generated Noisy-MNIST by adding periodic (every 10 pixels) white spots to the original MNIST test images, and Occluded-MNIST by putting a 5-pixel diameter dark circle at position (14,20) of the original MNIST test images.

For the evaluation of the dirty case discussed in Section 3.2.2, we have mixed global anomaly samples in the training phase by swapping 50 MNIST samples with 50 Fashion-MNIST samples. In this configuration, the ratio of global anomaly samples in the training dataset is about 1\%.

\begin{table}[!t]
\caption{Configuration of the autoencoder used in the evaluation. $(i, o, k, p, op)$ stands for ($input$, $output$, $kernel\_size$, $padding$, $output\_padding$). We have used $stride=2$ in Conv2D and TransConv2D.}
\begin{center}
    \begin{tabular}{|l|l|}
        \hline
        \textbf{Comp.} & \textbf{Layer (Parameters) } \\
        \hline
        Encoder & Conv2D ($i$=1, $o$=8, $k$=3, $p$=1) - ReLU -\\ 
        & Conv2D ($i$=8, $o$=16, $k$=3, $p$=1) -\\
        & BatchNorm2d (16) - ReLU -\\
        & Conv2D ($i$=16, $o$=32, $k$=3, $p$=0) - Flatten - \\ 
        & ReLU - Linear ($i$=288, $o$=128) - \\
        & ReLU - Linear ($i$=128, $o$=64) \\
    
        \hline
        Decoder &
        Linear ($i$=64, $o$=128) - ReLU - \\
        &Linear ($i$=128, $o$=288) - ReLU - Unflatten -\\
        & TransConv2D ($i$=32, $o$=16, $k$=3, $p$=0, $op$=0) - \\      
        & BatchNorm2d (16) - ReLU -\\
        & TransConv2D ($i$=16, $o$=8, $k$=3, $p$=1, $op$=1) - \\      
        & BatchNorm2d (8) - ReLU - \\
        & TransConv2D ($i$=8, $o$=1, $k$=3, $p$=1, $op$=1) - \\      
        & Sigmoid \\
        \hline
    \end{tabular}
\label{tab:ae_parameters}
\end{center}
\end{table}

We have used random waypoint mobility to generate the contacts of devices, which is very common in the studies of WANET \cite{bettstetter2004stochastic}. We have used RWP0500 -- the most representative mobility pattern in \cite{ochiai2022wireless}. According to the work, there are no huge differences in the performance among other mobility patterns. Of course, there can be a discussion of some special cases that a device is isolated from all other devices all the time. But, such physical failure cases are not our target.

\begin{figure}
\centering

  \begin{minipage}[b]{0.44\linewidth}
  \centering
  \includegraphics[keepaspectratio, scale=0.124]{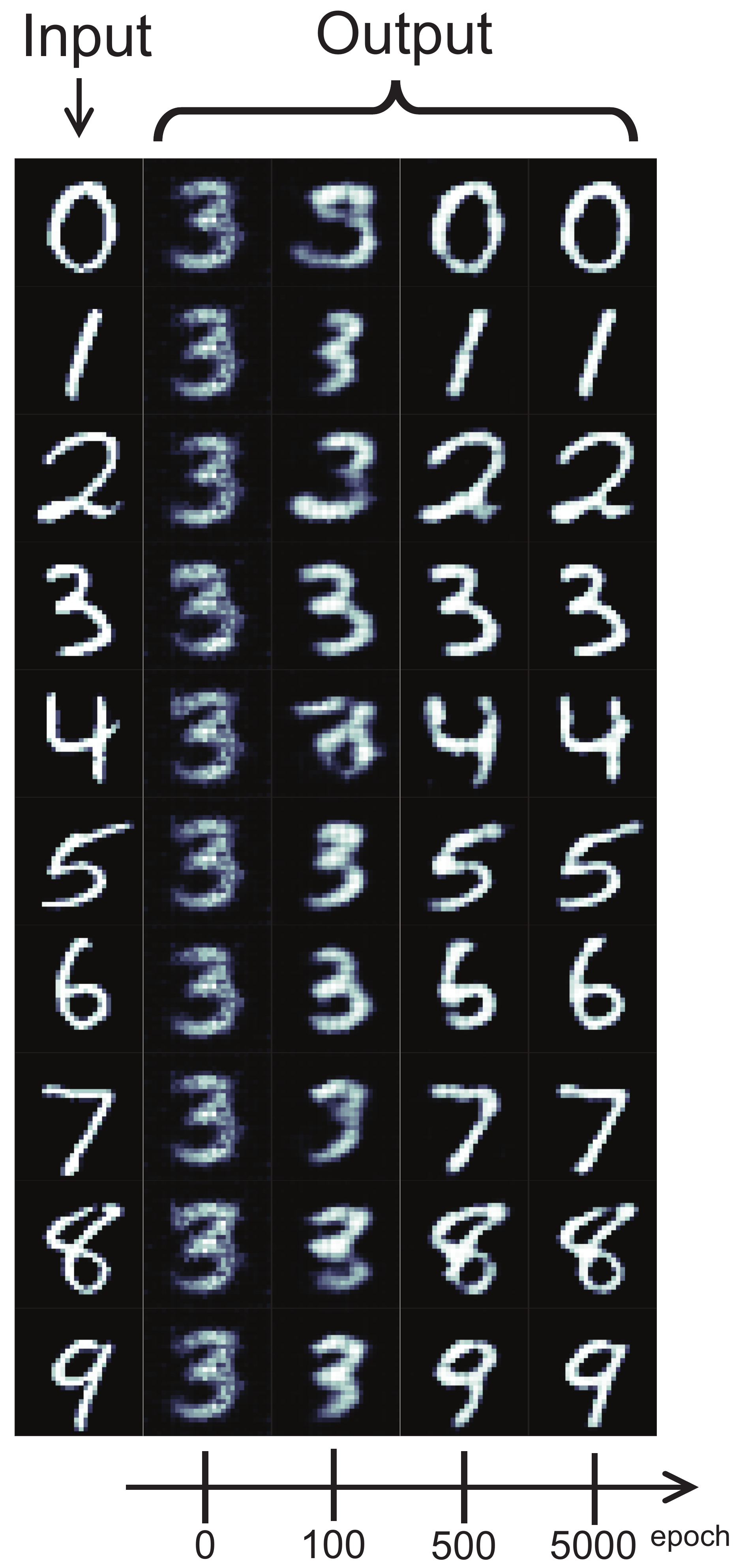}\\
  (a) Legitimate Inputs
  \end{minipage}
  \begin{minipage}[b]{0.44\linewidth}
  \centering
  \includegraphics[keepaspectratio, scale=0.124]{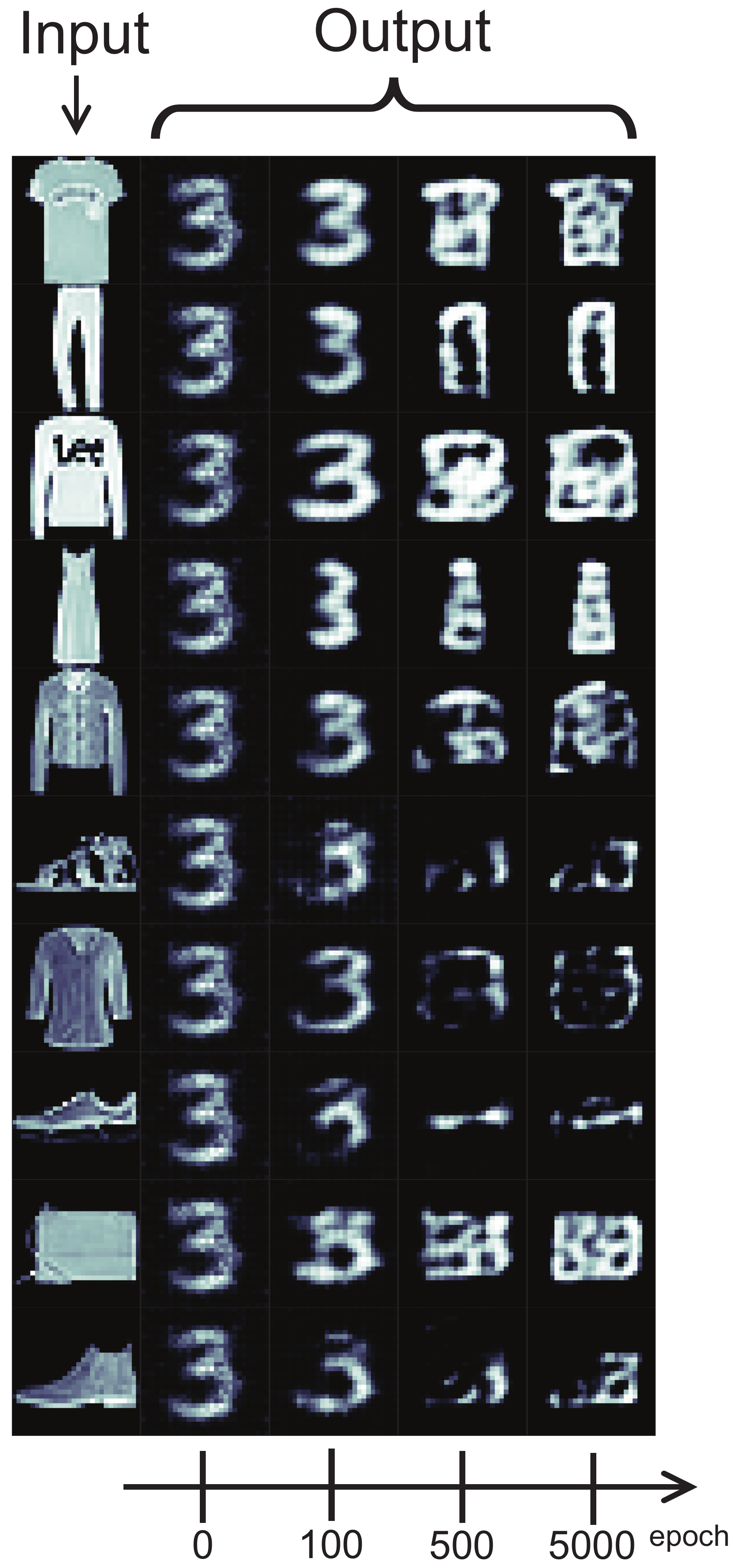}\\
  (b) Global Anomaly Inputs
  \end{minipage}
\caption{Reconstructed images of (a) legitimate inputs and (b) global anomaly inputs at epoch = 0, 100, 500, and 5000 (at device 3). The model could reconstruct minor legitimate class images successfully while keeping global anomaly outputs unconstructed.}
\label{fig:Reconstruction1}
\end{figure}

Table \ref{tab:ae_parameters} shows the detailed configuration of the autoencoder used in our experiment. For aggregation, model parameters except for batch normalization layers were exchanged among the encountered devices. 

We have used an SGD optimizer with momentum $=$ 0.9. Other hyper-parameters were: learning rate $=$ 0.001, batch size $=$ 32, coefficient of WAFL-aggregation $=$ 0.1, coefficient of threshold aggregation $\gamma=0.01$, and threshold ratio $\delta=0.999$. 

\subsection{Reconstructed Images}

Fig. \ref{fig:Reconstruction1} shows the reconstructed examples of (a) MNIST and (b) Fashion-MNIST test samples. At the early stage, e.g., at epochs 0 and 100, it could not reconstruct the images well except for the major classes of training data, which is '3' as observed. 

After several interactions among the devices, i.e., after epoch 500, they could reconstruct the images very well for all the MNIST classes in Fig. \ref{fig:Reconstruction1}(a). They did not reconstruct the images in the case of Fashion-MNIST in Fig. \ref{fig:Reconstruction1}(b), which is successful for global anomaly detection.

This result indicates that general legitimate features were trained successfully across the partitioned data even with the severe 99.95\% Non-IID case whereas global anomalies can be detected by comparing the inputs and the reconstructions.

\subsection{Detection of Global Anomalies}

\begin{figure}
\centering
\includegraphics[width=0.43\textwidth]{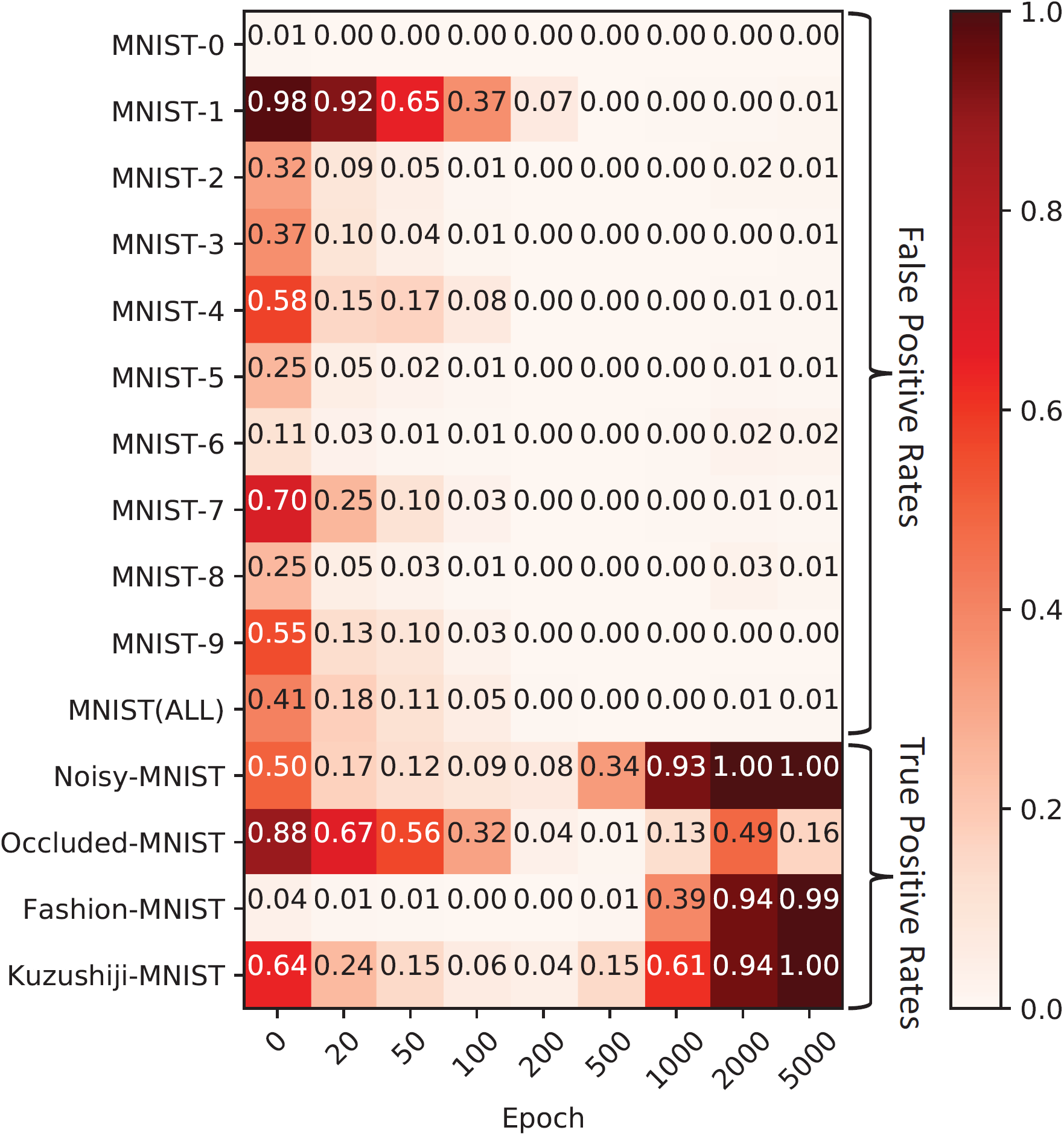}
\caption{Positive rates of anomaly detection to the test data at device 0 for the models from epoch 0 to 5000. MNIST are false positive rates (FPR). Noisy-, Occluded-, Fashion-, and Kuzhushiji-MNIST are true positive rates (TPR). The number $y$ in MNIST-$y$ indicates FPR for class $y$ test samples. }
\label{fig:DetectionRate}
\end{figure}

Fig. \ref{fig:DetectionRate} shows the trend of positive rates for the detection by class or by category of the test data from epoch 0 to epoch 5000 at device 0 in the pure case. The number after MNIST (e.g., MNIST-0 $\ldots$ MNIST-9) indicates the class. MNIST(ALL) means all the classes are mixed together. As MNIST is legitimate in the global anomaly context, the rates of MNIST are false positive rates (FPR). For Noisy-, Occluded-, Fashion-, and Kuzhushiji-MNIST are the intentional global anomaly samples, the rates of them are true positive rates (TPR). In this experiment, we have used three instances of RWP mobilities with the same hyper-parameters but with different random seeds. The FPRs and TPRs are the averages of them.

\begin{table*}[t]
\centering
\caption{True/False positive rates by the developed models and thresholds at epoch 5000 per node. N, O, F, and K indicate Noisy-, Occluded-, Fashion-, and Kuzushiji-MNIST. Higher TPR with lower FPR is better. \textit{Self-train fails in achieving low FPRs.}\label{tab:tfpr_summary}}
\begin{tabular}{|c||ccccc||ccccc||ccccc|}
\hline
\multirow{3}{*}{\textbf{Device}}&\multicolumn{5}{|c||}{\textbf{WAFL-AE (train without anomaly)}}&\multicolumn{5}{|c||}{\textbf{WAFL-AE (train with 1\% anomaly)}}&\multicolumn{5}{|c|}{\textbf{Self-train (without anomaly)}} \\
\cline{2-16}
&\multicolumn{4}{c}{TPR}&\multicolumn{1}{|c||}{\multirow{2}{*}{FPR}}&\multicolumn{4}{c}{TPR}&\multicolumn{1}{|c||}{\multirow{2}{*}{FPR}}&\multicolumn{4}{c}{TPR}&\multicolumn{1}{|c|}{\multirow{2}{*}{FPR}} \\ 
\cline{2-5}\cline{7-10}\cline{12-15}
&N&O&F&K&\multicolumn{1}{|c||}{}&N&O&F&K&\multicolumn{1}{|c||}{}&N&O&F&K&\multicolumn{1}{|c|}{} \\ \hline
\textbf{0}&100\%&16.2\%&99.5\%&99.5\%&\multicolumn{1}{|c||}{0.68\%} &100\%&15.9\%&92.3\%&99.1\%&\multicolumn{1}{|c||}{0.54\%} &100\%&84.3\%&98.1\%&99.6\%&\multicolumn{1}{|c|}{42.2\%} \\
\textbf{1}&100\%&5.75\%&100\%&100\%&\multicolumn{1}{|c||}{3.66\%} &100\%&4.53\%&100\%&100\%&\multicolumn{1}{|c||}{3.27\%} &100\%&80.7\%&98.7\%&99.6\%&\multicolumn{1}{|c|}{63.5\%}\\
\textbf{2}&100\%&5.21\%&99.5\%&99.6\%&\multicolumn{1}{|c||}{0.33\%} &100\%&5.16\%&97.3\%&99.7\%&\multicolumn{1}{|c||}{0.32\%} &100\%&25.3\%&97.1\%&97.7\%&\multicolumn{1}{|c|}{3.06\%}\\
\textbf{3}&100\%&5.60\%&99.6\%&99.5\%&\multicolumn{1}{|c||}{0.40\%} &100\%&5.94\%&97.9\%&99.7\%&\multicolumn{1}{|c||}{0.43\%} &100\%&18.3\%&97.8\%&96.4\%&\multicolumn{1}{|c|}{2.50\%}\\
\textbf{4}&100\%&3.09\%&99.8\%&99.8\%&\multicolumn{1}{|c||}{0.39\%} &100\%&3.00\%&99.6\%&99.9\%&\multicolumn{1}{|c||}{0.37\%} &100\%&39.6\%&99.6\%&99.7\%&\multicolumn{1}{|c|}{22.1\%}\\
\textbf{5}&100\%&2.82\%&99.7\%&99.7\%&\multicolumn{1}{|c||}{0.33\%} &100\%&2.74\%&99.6\%&99.9\%&\multicolumn{1}{|c||}{0.32\%} &100\%&20.9\%&99.4\%&99.1\%&\multicolumn{1}{|c|}{3.54\%}\\
\textbf{6}&100\%&9.97\%&99.8\%&99.8\%&\multicolumn{1}{|c||}{1.22\%} &100\%&10.8\%&98.7\%&99.9\%&\multicolumn{1}{|c||}{0.97\%} &99.8\%&52.9\%&94.1\%&98.8\%&\multicolumn{1}{|c|}{24.4\%}\\
\textbf{7}&100\%&2.43\%&99.9\%&99.9\%&\multicolumn{1}{|c||}{0.48\%} &100\%&2.36\%&99.3\%&99.9\%&\multicolumn{1}{|c||}{0.46\%} &100\%&41.8\%&96.7\%&98.4\%&\multicolumn{1}{|c|}{16.2\%}\\
\textbf{8}&100\%&8.97\%&99.7\%&99.6\%&\multicolumn{1}{|c||}{0.38\%} &100\%&7.75\%&98.0\%&99.8\%&\multicolumn{1}{|c||}{0.38\%} &99.9\%&9.01\%&90.2\%&94.4\%&\multicolumn{1}{|c|}{1.72\%}\\
\textbf{9}&100\%&3.69\%&99.9\%&99.9\%&\multicolumn{1}{|c||}{0.40\%} &100\%&3.27\%&99.4\%&99.9\%&\multicolumn{1}{|c||}{0.36\%} &100\%&24.9\%&99.3\%&98.8\%&\multicolumn{1}{|c|}{8.58\%}\\
\hline
\textbf{Avg.}&100\%&6.37\%&99.7\%&99.7\%&\multicolumn{1}{|c||}{0.83\%} &100\%&6.12\%&98.2\%&99.8\%&\multicolumn{1}{|c||}{0.74\%} &100\%&40.0\%&97.1\%&98.2\%&\multicolumn{1}{|c|}{18.8\%} \\ \hline
\end{tabular}
\end{table*}

Please note that the major training data at device 0 are MNIST class 0, and it achieved good (i.e., low) FPR for MNIST-0 all the time even from epoch 0. Other MNIST classes had much higher FPR at the beginning, but it almost reached zero from epoch 200. This indicates that the WAFL-Autoencoder has worked effectively for learning other legitimate data samples beyond the major class of the device.

As for global anomalies, device 0 started to detect from epoch 500, finally reaching the TPRs of 100\%, 16\%, 99\%, and 100\% for Noisy-, Occluded-, Fashion-, and Kuzushiji-MNIST respectively at 5000. 

We have observed two steps of stabilization for anomaly detection. The first step was model stabilization. It took the first 1000 epochs for generalizing the autoencoder model itself. The next step was threshold stabilization. After stabilizing the model, the threshold could be stabilized.

Table \ref{tab:tfpr_summary} shows the summary of True / False positive rates by the developed models and thresholds per device at epoch 5000 under (1) train without anomaly, (2) train with 1\% anomaly samples, and (3) self-training without anomaly as a baseline. Compared to the self-training, WAFL-Autoencoder achieved good performance on average showing lower false positive rates and higher positive rates except for Occluded-MNIST in both pure and dirty cases.

\section{Conclusion}

We have proposed the WAFL-Autoencoder that trains distributed autoencoders with device-to-device communications assuming the application of multiple IoT devices physically located nearby. We have also proposed a distributed threshold-finding algorithm for finding Global anomalies with WAFL-Autoencoder.

With our standard benchmark-based evaluation, we have confirmed that WAFL-Autoencoder trains autoencoders perfectly for reconstructing the inputs of legitimate data across the devices. We have also confirmed that the devices collaboratively found thresholds for global anomaly detection with low false positive rates while achieving high true positive rates for global anomalies with few exceptions.

Future work may include the application of more realistic data such as electric monitoring sequences, accelerometer signals, and photo images.

\section*{Acknowledgement}
This paper is based on the results obtained from a project
commissioned by the New Energy and Industrial Technology
Development Organization (NEDO), Japan. This work was also supported by JSPS KAKENHI Grant Number JP 22H03572.

\bibliographystyle{ACM-Reference-Format} 
\bibliography{acmart}


\begin{thebibliography}{29}


\ifx \showCODEN    \undefined \def \showCODEN     #1{\unskip}     \fi
\ifx \showDOI      \undefined \def \showDOI       #1{#1}\fi
\ifx \showISBNx    \undefined \def \showISBNx     #1{\unskip}     \fi
\ifx \showISBNxiii \undefined \def \showISBNxiii  #1{\unskip}     \fi
\ifx \showISSN     \undefined \def \showISSN      #1{\unskip}     \fi
\ifx \showLCCN     \undefined \def \showLCCN      #1{\unskip}     \fi
\ifx \shownote     \undefined \def \shownote      #1{#1}          \fi
\ifx \showarticletitle \undefined \def \showarticletitle #1{#1}   \fi
\ifx \showURL      \undefined \def \showURL       {\relax}        \fi
\providecommand\bibfield[2]{#2}
\providecommand\bibinfo[2]{#2}
\providecommand\natexlab[1]{#1}
\providecommand\showeprint[2][]{arXiv:#2}

\bibitem[Abolhasan et~al\mbox{.}(2004)]%
        {abolhasan2004review}
\bibfield{author}{\bibinfo{person}{Mehran Abolhasan}, \bibinfo{person}{Tadeusz
  Wysocki}, {and} \bibinfo{person}{Eryk Dutkiewicz}.}
  \bibinfo{year}{2004}\natexlab{}.
\newblock \showarticletitle{A review of routing protocols for mobile ad hoc
  networks}.
\newblock \bibinfo{journal}{\emph{Ad hoc networks}} \bibinfo{volume}{2},
  \bibinfo{number}{1} (\bibinfo{year}{2004}), \bibinfo{pages}{1--22}.
\newblock


\bibitem[Akcay et~al\mbox{.}(2018)]%
        {akcay2018ganomaly}
\bibfield{author}{\bibinfo{person}{Samet Akcay}, \bibinfo{person}{Amir
  Atapour-Abarghouei}, {and} \bibinfo{person}{Toby~P Breckon}.}
  \bibinfo{year}{2018}\natexlab{}.
\newblock \showarticletitle{Ganomaly: Semi-supervised anomaly detection via
  adversarial training}. In \bibinfo{booktitle}{\emph{Asian conference on
  computer vision}}. Springer, \bibinfo{pages}{622--637}.
\newblock


\bibitem[Anaissi et~al\mbox{.}(2022)]%
        {anaissi2022personalized}
\bibfield{author}{\bibinfo{person}{Ali Anaissi}, \bibinfo{person}{Basem
  Suleiman}, {and} \bibinfo{person}{Widad Alyassine}.}
  \bibinfo{year}{2022}\natexlab{}.
\newblock \showarticletitle{A Personalized Federated Learning Algorithm for
  One-Class Support Vector Machine: An Application in Anomaly Detection}. In
  \bibinfo{booktitle}{\emph{International Conference on Computational
  Science}}. Springer, \bibinfo{pages}{373--379}.
\newblock


\bibitem[Bao et~al\mbox{.}(2019)]%
        {bao2019flchain}
\bibfield{author}{\bibinfo{person}{Xianglin Bao}, \bibinfo{person}{Cheng Su},
  \bibinfo{person}{Yan Xiong}, \bibinfo{person}{Wenchao Huang}, {and}
  \bibinfo{person}{Yifei Hu}.} \bibinfo{year}{2019}\natexlab{}.
\newblock \showarticletitle{{FLchain}: A blockchain for auditable federated
  learning with trust and incentive}. In \bibinfo{booktitle}{\emph{2019 5th
  International Conference on Big Data Computing and Communications (BIGCOM)}}.
  IEEE, \bibinfo{pages}{151--159}.
\newblock


\bibitem[Bettstetter et~al\mbox{.}(2004)]%
        {bettstetter2004stochastic}
\bibfield{author}{\bibinfo{person}{Christian Bettstetter},
  \bibinfo{person}{Hannes Hartenstein}, {and} \bibinfo{person}{Xavier
  P{\'e}rez-Costa}.} \bibinfo{year}{2004}\natexlab{}.
\newblock \showarticletitle{Stochastic properties of the random waypoint
  mobility model}.
\newblock \bibinfo{journal}{\emph{Wireless networks}} \bibinfo{volume}{10},
  \bibinfo{number}{5} (\bibinfo{year}{2004}), \bibinfo{pages}{555--567}.
\newblock


\bibitem[Clanuwat et~al\mbox{.}(2018)]%
        {clanuwat2018deep}
\bibfield{author}{\bibinfo{person}{Tarin Clanuwat}, \bibinfo{person}{Mikel
  Bober-Irizar}, \bibinfo{person}{Asanobu Kitamoto}, \bibinfo{person}{Alex
  Lamb}, \bibinfo{person}{Kazuaki Yamamoto}, {and} \bibinfo{person}{David Ha}.}
  \bibinfo{year}{2018}\natexlab{}.
\newblock \showarticletitle{Deep learning for classical japanese literature}.
\newblock \bibinfo{journal}{\emph{arXiv preprint arXiv:1812.01718}}
  (\bibinfo{year}{2018}).
\newblock


\bibitem[Cui et~al\mbox{.}(2021)]%
        {cui2021security}
\bibfield{author}{\bibinfo{person}{Lei Cui}, \bibinfo{person}{Youyang Qu},
  \bibinfo{person}{Gang Xie}, \bibinfo{person}{Deze Zeng},
  \bibinfo{person}{Ruidong Li}, \bibinfo{person}{Shigen Shen}, {and}
  \bibinfo{person}{Shui Yu}.} \bibinfo{year}{2021}\natexlab{}.
\newblock \showarticletitle{Security and privacy-enhanced federated learning
  for anomaly detection in iot infrastructures}.
\newblock \bibinfo{journal}{\emph{IEEE Transactions on Industrial Informatics}}
  \bibinfo{volume}{18}, \bibinfo{number}{5} (\bibinfo{year}{2021}),
  \bibinfo{pages}{3492--3500}.
\newblock


\bibitem[Fall(2003)]%
        {fall2003delay}
\bibfield{author}{\bibinfo{person}{Kevin Fall}.}
  \bibinfo{year}{2003}\natexlab{}.
\newblock \showarticletitle{A delay-tolerant network architecture for
  challenged {Internets}}. In \bibinfo{booktitle}{\emph{Proceedings of the 2003
  conference on Applications, technologies, architectures, and protocols for
  computer communications}}. \bibinfo{pages}{27--34}.
\newblock


\bibitem[Frodigh et~al\mbox{.}(2000)]%
        {frodigh2000wireless}
\bibfield{author}{\bibinfo{person}{Magnus Frodigh}, \bibinfo{person}{Per
  Johansson}, {and} \bibinfo{person}{Peter Larsson}.}
  \bibinfo{year}{2000}\natexlab{}.
\newblock \showarticletitle{Wireless ad hoc networking: the art of networking
  without a network}.
\newblock \bibinfo{journal}{\emph{Ericsson review}} \bibinfo{volume}{4},
  \bibinfo{number}{4} (\bibinfo{year}{2000}), \bibinfo{pages}{249}.
\newblock


\bibitem[Hartenstein and Laberteaux(2008)]%
        {hartenstein2008tutorial}
\bibfield{author}{\bibinfo{person}{Hannes Hartenstein} {and}
  \bibinfo{person}{LP Laberteaux}.} \bibinfo{year}{2008}\natexlab{}.
\newblock \showarticletitle{A tutorial survey on vehicular ad hoc networks}.
\newblock \bibinfo{journal}{\emph{IEEE Communications magazine}}
  \bibinfo{volume}{46}, \bibinfo{number}{6} (\bibinfo{year}{2008}),
  \bibinfo{pages}{164--171}.
\newblock


\bibitem[Hu et~al\mbox{.}(2019)]%
        {hu2019decentralized}
\bibfield{author}{\bibinfo{person}{Chenghao Hu}, \bibinfo{person}{Jingyan
  Jiang}, {and} \bibinfo{person}{Zhi Wang}.} \bibinfo{year}{2019}\natexlab{}.
\newblock \showarticletitle{Decentralized federated learning: A segmented
  gossip approach}.
\newblock \bibinfo{journal}{\emph{arXiv preprint arXiv:1908.07782}}
  (\bibinfo{year}{2019}).
\newblock


\bibitem[Kone{\v{c}}n{\`y} et~al\mbox{.}(2016)]%
        {konevcny2016federated}
\bibfield{author}{\bibinfo{person}{Jakub Kone{\v{c}}n{\`y}},
  \bibinfo{person}{H~Brendan McMahan}, \bibinfo{person}{Felix~X Yu},
  \bibinfo{person}{Peter Richt{\'a}rik}, \bibinfo{person}{Ananda~Theertha
  Suresh}, {and} \bibinfo{person}{Dave Bacon}.}
  \bibinfo{year}{2016}\natexlab{}.
\newblock \showarticletitle{Federated learning: Strategies for improving
  communication efficiency}.
\newblock \bibinfo{journal}{\emph{arXiv preprint arXiv:1610.05492}}
  (\bibinfo{year}{2016}).
\newblock


\bibitem[Li et~al\mbox{.}(2020b)]%
        {li2020federated}
\bibfield{author}{\bibinfo{person}{Tian Li}, \bibinfo{person}{Anit~Kumar Sahu},
  \bibinfo{person}{Ameet Talwalkar}, {and} \bibinfo{person}{Virginia Smith}.}
  \bibinfo{year}{2020}\natexlab{b}.
\newblock \showarticletitle{Federated learning: Challenges, methods, and future
  directions}.
\newblock \bibinfo{journal}{\emph{IEEE Signal Processing Magazine}}
  \bibinfo{volume}{37}, \bibinfo{number}{3} (\bibinfo{year}{2020}),
  \bibinfo{pages}{50--60}.
\newblock


\bibitem[Li et~al\mbox{.}(2020a)]%
        {li2020blockchain}
\bibfield{author}{\bibinfo{person}{Yuzheng Li}, \bibinfo{person}{Chuan Chen},
  \bibinfo{person}{Nan Liu}, \bibinfo{person}{Huawei Huang},
  \bibinfo{person}{Zibin Zheng}, {and} \bibinfo{person}{Qiang Yan}.}
  \bibinfo{year}{2020}\natexlab{a}.
\newblock \showarticletitle{A blockchain-based decentralized federated learning
  framework with committee consensus}.
\newblock \bibinfo{journal}{\emph{IEEE Network}} \bibinfo{volume}{35},
  \bibinfo{number}{1} (\bibinfo{year}{2020}), \bibinfo{pages}{234--241}.
\newblock


\bibitem[Martin(2021)]%
        {martin2021fault}
\bibfield{author}{\bibinfo{person}{Damien~W Martin}.}
  \bibinfo{year}{2021}\natexlab{}.
\newblock \emph{\bibinfo{title}{Fault detection in manufacturing equipment
  using unsupervised deep learning}}.
\newblock \bibinfo{thesistype}{Ph.\,D. Dissertation}.
  \bibinfo{school}{Massachusetts Institute of Technology}.
\newblock


\bibitem[Ochiai et~al\mbox{.}(2022)]%
        {ochiai2022wireless}
\bibfield{author}{\bibinfo{person}{Hideya Ochiai}, \bibinfo{person}{Yuwei Sun},
  \bibinfo{person}{Qingzhe Jin}, \bibinfo{person}{Nattanon Wongwiwatchai},
  {and} \bibinfo{person}{Hiroshi Esaki}.} \bibinfo{year}{2022}\natexlab{}.
\newblock \showarticletitle{Wireless Ad Hoc Federated Learning: A Fully
  Distributed Cooperative Machine Learning}.
\newblock \bibinfo{journal}{\emph{arXiv preprint arXiv:2205.11779}}
  (\bibinfo{year}{2022}).
\newblock


\bibitem[Palato(2021)]%
        {palato2021federated}
\bibfield{author}{\bibinfo{person}{Mirko Palato}.}
  \bibinfo{year}{2021}\natexlab{}.
\newblock \showarticletitle{Federated variational autoencoder for collaborative
  filtering}. In \bibinfo{booktitle}{\emph{2021 International Joint Conference
  on Neural Networks (IJCNN)}}. IEEE, \bibinfo{pages}{1--8}.
\newblock


\bibitem[Pappas et~al\mbox{.}(2021)]%
        {pappas2021ipls}
\bibfield{author}{\bibinfo{person}{Christodoulos Pappas},
  \bibinfo{person}{Dimitris Chatzopoulos}, \bibinfo{person}{Spyros Lalis},
  {and} \bibinfo{person}{Manolis Vavalis}.} \bibinfo{year}{2021}\natexlab{}.
\newblock \showarticletitle{{IPLS}: A framework for decentralized federated
  learning}. In \bibinfo{booktitle}{\emph{2021 IFIP Networking Conference}}.
  IEEE, \bibinfo{pages}{1--6}.
\newblock


\bibitem[Pei et~al\mbox{.}(2022)]%
        {pei2022personalized}
\bibfield{author}{\bibinfo{person}{Jiaming Pei}, \bibinfo{person}{Kaiyang
  Zhong}, \bibinfo{person}{Mian~Ahmad Jan}, {and} \bibinfo{person}{Jinhai Li}.}
  \bibinfo{year}{2022}\natexlab{}.
\newblock \showarticletitle{Personalized federated learning framework for
  network traffic anomaly detection}.
\newblock \bibinfo{journal}{\emph{Computer Networks}}  \bibinfo{volume}{209}
  (\bibinfo{year}{2022}), \bibinfo{pages}{108906}.
\newblock


\bibitem[Pokhrel and Choi(2020)]%
        {pokhrel2020federated}
\bibfield{author}{\bibinfo{person}{Shiva~Raj Pokhrel} {and}
  \bibinfo{person}{Jinho Choi}.} \bibinfo{year}{2020}\natexlab{}.
\newblock \showarticletitle{Federated learning with blockchain for autonomous
  vehicles: Analysis and design challenges}.
\newblock \bibinfo{journal}{\emph{IEEE Transactions on Communications}}
  \bibinfo{volume}{68}, \bibinfo{number}{8} (\bibinfo{year}{2020}),
  \bibinfo{pages}{4734--4746}.
\newblock


\bibitem[Roy et~al\mbox{.}(2019)]%
        {roy2019braintorrent}
\bibfield{author}{\bibinfo{person}{Abhijit~Guha Roy}, \bibinfo{person}{Shayan
  Siddiqui}, \bibinfo{person}{Sebastian P{\"o}lsterl}, \bibinfo{person}{Nassir
  Navab}, {and} \bibinfo{person}{Christian Wachinger}.}
  \bibinfo{year}{2019}\natexlab{}.
\newblock \showarticletitle{{BrainTorrent}: A peer-to-peer environment for
  decentralized federated learning}.
\newblock \bibinfo{journal}{\emph{arXiv preprint arXiv:1905.06731}}
  (\bibinfo{year}{2019}).
\newblock


\bibitem[Sater and Hamza(2021)]%
        {sater2021federated}
\bibfield{author}{\bibinfo{person}{Raed~Abdel Sater} {and}
  \bibinfo{person}{A~Ben Hamza}.} \bibinfo{year}{2021}\natexlab{}.
\newblock \showarticletitle{A federated learning approach to anomaly detection
  in smart buildings}.
\newblock \bibinfo{journal}{\emph{ACM Transactions on Internet of Things}}
  \bibinfo{volume}{2}, \bibinfo{number}{4} (\bibinfo{year}{2021}),
  \bibinfo{pages}{1--23}.
\newblock


\bibitem[Tabassum et~al\mbox{.}(2022)]%
        {tabassum2022fedgan}
\bibfield{author}{\bibinfo{person}{Aliya Tabassum}, \bibinfo{person}{Aiman
  Erbad}, \bibinfo{person}{Wadha Lebda}, \bibinfo{person}{Amr Mohamed}, {and}
  \bibinfo{person}{Mohsen Guizani}.} \bibinfo{year}{2022}\natexlab{}.
\newblock \showarticletitle{FEDGAN-IDS: Privacy-preserving IDS using GAN and
  Federated Learning}.
\newblock \bibinfo{journal}{\emph{Computer Communications}}
  (\bibinfo{year}{2022}).
\newblock


\bibitem[Tomiyama et~al\mbox{.}(2023)]%
        {tomiyama2023gan}
\bibfield{author}{\bibinfo{person}{Eisuke Tomiyama}, \bibinfo{person}{Hiroshi
  Esaki}, {and} \bibinfo{person}{Hideya Ochiai}.}
  \bibinfo{year}{2023}\natexlab{}.
\newblock \showarticletitle{{WAFL-GAN:} Wireless Ad Hoc Federated Learning for
  Distributed Generative Adversarial Networks}. In
  \bibinfo{booktitle}{\emph{IEEE International Conference on Knowledge and
  Smart Technology}}.
\newblock


\bibitem[ur~Rehman et~al\mbox{.}(2020)]%
        {ur2020towards}
\bibfield{author}{\bibinfo{person}{Muhammad~Habib ur Rehman},
  \bibinfo{person}{Khaled Salah}, \bibinfo{person}{Ernesto Damiani}, {and}
  \bibinfo{person}{Davor Svetinovic}.} \bibinfo{year}{2020}\natexlab{}.
\newblock \showarticletitle{Towards blockchain-based reputation-aware federated
  learning}. In \bibinfo{booktitle}{\emph{IEEE Conference on Computer
  Communications Workshops (INFOCOM WKSHPS)}}. IEEE, \bibinfo{pages}{183--188}.
\newblock


\bibitem[Warnat-Herresthal et~al\mbox{.}(2021)]%
        {warnat2021swarm}
\bibfield{author}{\bibinfo{person}{Stefanie Warnat-Herresthal},
  \bibinfo{person}{Hartmut Schultze},
  \bibinfo{person}{Krishnaprasad~Lingadahalli Shastry},
  \bibinfo{person}{Sathyanarayanan Manamohan}, {et~al\mbox{.}}}
  \bibinfo{year}{2021}\natexlab{}.
\newblock \showarticletitle{Swarm learning for decentralized and confidential
  clinical machine learning}.
\newblock \bibinfo{journal}{\emph{Nature}} \bibinfo{volume}{594},
  \bibinfo{number}{7862} (\bibinfo{year}{2021}), \bibinfo{pages}{265--270}.
\newblock


\bibitem[Weinger et~al\mbox{.}(2022)]%
        {weinger2022enhancing}
\bibfield{author}{\bibinfo{person}{Brett Weinger}, \bibinfo{person}{Jinoh Kim},
  \bibinfo{person}{Alex Sim}, \bibinfo{person}{Makiya Nakashima},
  \bibinfo{person}{Nour Moustafa}, {and} \bibinfo{person}{K~John Wu}.}
  \bibinfo{year}{2022}\natexlab{}.
\newblock \showarticletitle{Enhancing IoT anomaly detection performance for
  federated learning}.
\newblock \bibinfo{journal}{\emph{Digital Communications and Networks}}
  (\bibinfo{year}{2022}).
\newblock


\bibitem[Zhang et~al\mbox{.}(2021)]%
        {zhang2021federated}
\bibfield{author}{\bibinfo{person}{Kai Zhang}, \bibinfo{person}{Yushan Jiang},
  \bibinfo{person}{Lee Seversky}, \bibinfo{person}{Chengtao Xu},
  \bibinfo{person}{Dahai Liu}, {and} \bibinfo{person}{Houbing Song}.}
  \bibinfo{year}{2021}\natexlab{}.
\newblock \showarticletitle{Federated variational learning for anomaly
  detection in multivariate time series}. In \bibinfo{booktitle}{\emph{2021
  IEEE International Performance, Computing, and Communications Conference
  (IPCCC)}}. IEEE, \bibinfo{pages}{1--9}.
\newblock


\bibitem[Zhao et~al\mbox{.}(2019)]%
        {zhao2019multi}
\bibfield{author}{\bibinfo{person}{Ying Zhao}, \bibinfo{person}{Junjun Chen},
  \bibinfo{person}{Di Wu}, \bibinfo{person}{Jian Teng}, {and}
  \bibinfo{person}{Shui Yu}.} \bibinfo{year}{2019}\natexlab{}.
\newblock \showarticletitle{Multi-task network anomaly detection using
  federated learning}. In \bibinfo{booktitle}{\emph{Proceedings of the tenth
  international symposium on information and communication technology}}.
  \bibinfo{pages}{273--279}.
\newblock


\end{thebibliography}


\end{document}